\documentclass{article}
\usepackage{ijcai22}

\usepackage{times}

\usepackage{soul}
\usepackage{url}
\usepackage[hidelinks]{hyperref}
\usepackage[utf8]{inputenc}
\usepackage[small]{caption}
\usepackage{graphicx}
\usepackage{amsmath}
\usepackage{amsthm}
\usepackage{booktabs}
\usepackage{xspace} 
\usepackage{algorithm}  
\usepackage{algpseudocode}  
\usepackage{amsmath}  
\usepackage{booktabs}
\usepackage{float}

\usepackage{stfloats}

\title{CorefDRE: Document-level Relation Extraction with coreference resolution}

\author{
Zhongxuan Xue$^1$\and
Jiang Zhong$^1$\and
Qizhu Dai$^1$\And
Rongzhen Li$^1$\footnote{Contact Author}\\\
\affiliations
$^1$Chongqing University\\
\emails
\{20164351, zhongjiang, daiqizhu, lirongzhen\}@cqu.edu.cn
}

\begin{document}

\maketitle

\begin{abstract}
Document-level relation extraction is to extract relation facts from a document consisting of multiple sentences, in which pronoun crossed sentences are a ubiquitous phenomenon against a single sentence. However, most of the previous works focus more on mentions coreference resolution except for pronouns, and rarely pay attention to mention-pronoun coreference and capturing the relations. To represent multi-sentence features by pronouns, we imitate the reading process of humans by leveraging coreference information when dynamically constructing a heterogeneous graph to enhance semantic information. Since the pronoun is notoriously ambiguous in the graph, a mention-pronoun coreference resolution is introduced to calculate the affinity between pronouns and corresponding mentions, and the noise suppression mechanism is proposed to reduce the noise caused by pronouns. Experiments on the public dataset, DocRED, DialogRE and MPDD, show that Coref-aware Doc-level Relation Extraction based on Graph Inference Network outperforms the state-of-the-art \footnote{We will publicly release our source code.}.
\end{abstract}
\section{Introduction}

Relation Extraction (RE), a task that automatically extracts relation facts between two entities in a given text, is widely used in the knowledge base population~\cite{zhang2017position} and the recommendation system~\cite{zhang2021causerec}. Existing research methods are mainly  responsible for the sentence-level RE, which aims to identify relations between an entity pair in a single sentence. However, large amounts of relation facts can only be extracted through multiple sentences, which cannot be achieved by sentence-level relation extraction. Therefore, researchers gradually pay more attention to document-level RE.

\begin{figure}[ht]
    \centering
    \setlength{\abovecaptionskip}{0cm}
    \setlength{\belowcaptionskip}{-0.4cm}
    \includegraphics[width=3.3in]{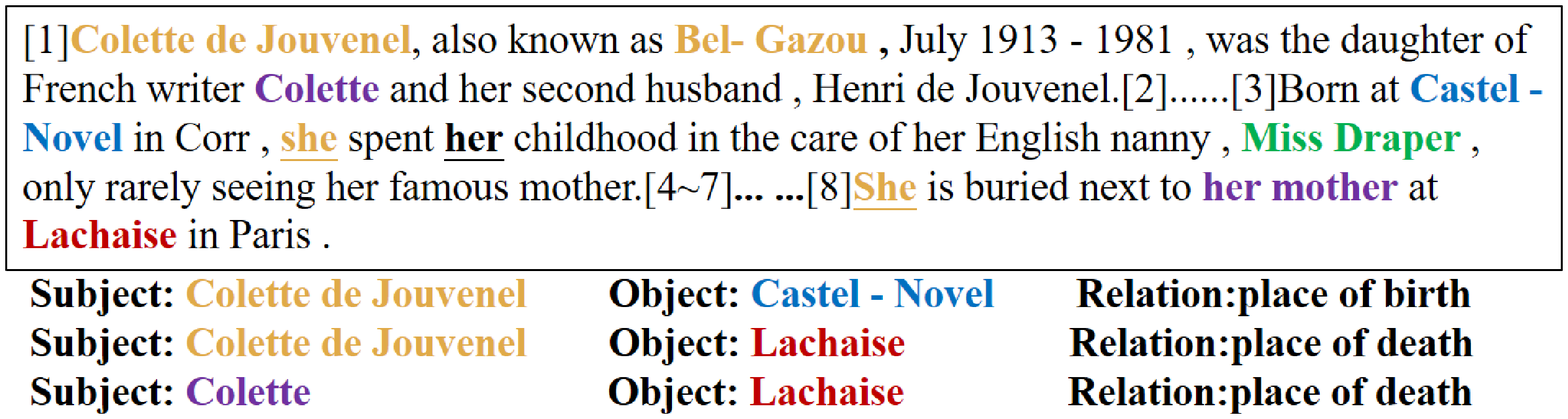}
    \caption{An example document from DocRED. Entities are distinguished by color, with the relation labels listed offside. $Colette$ $de$ $Jouvenel$ and $Bel$-$Gazou$ are general mentions coreference, while the pronoun crossed sentences $she$ point to mention $ Colette$ $de$ $Jouvenel$, which is a mention-pronoun coreference.}
    \label{fig1}
\end{figure}

The task of Doc-level RE needs to handle sentence-level relationships and capture complex interactions among cross-sentence entities in the document. Coreference resolution is required to dive deeply the relationship of entity pairs and the approach for reasoning neglects the interaction of mention and pronoun. Recent studies~\cite{ye2020coreferential,xu2021discriminative,huang2021graph} focus on graph-based reasoning skills, where coreference dependency, especially mention-mention coreference, is extensively used for logical inference. However, the pronouns, beneficial to multi-hop graph convolution, are ignored or used implicitly. The major challenge is how to explicitly model the relationship of mention-pronoun pairs for relation extraction crossing multi-sentences, while it is yet to be known whether modeling mention-pronoun coreference dependency is competitive with the intuitive reasoning based on the graph between subject entity and object entity.

Concretely, the graph-based method constructs the input document effectively but cannot explicitly capture the pronoun which is essential for Doc-level RE. Figure~\ref{fig1} shows an example from DocRED dataset \cite{yao2019docred}. $Colette$ $de$ $Jouvebel$ from the $1st$ sentence, $she$ and $her$ from the $3rd$ sentence and $she$ from the $8th$ sentence refer to the same entity, $Colette$ $de$ $Jouvebel$, only based on the facts, can we infer the relation fact of entity pair $\left< Colette~de~Jouvebel, Castel- Novel \right>$ is $place$ $of$ $birth$ and the relation fact of entity pair $\left< Colette~de~Jouvebel, Lachise \right>$ is $place$ $of$ $death$. In addition, only when we know that $her$ $mother$ in the $8th$ sentence refers to $colette$ in the $1st$ sentence can we infer that the relation fact between $colette$ and $Lachaise$ is $place$ $of$ $death$. Intuitively, the pronouns in document indicate rich semantic information, which is extremely vital to document-level RE. To verify the hypothesis, we randomly sample 100 documents from the DocRED training set and take stock of the pronouns and mention-pronoun pairs. Table~\ref{tab1} describes that pronouns, including $``you"$, $``your"$, $``yours"$, $``i"$, $``me"$, $``we"$, $``our"$, $``ours"$, $``he"$, $``him"$, $``his"$, $``she"$, $``her"$, $``they"$, $``their"$, $``them"$, $``it"$, $``its"$, are approximate to 31.69 per document and mention-pronoun pairs about 13.49. It is clear that pronouns can provide an important clue to Doc-level RE if some strategies are designed ingeniously.

\begin{table}[ht]
    \centering
     \setlength{\belowcaptionskip}{-0.2cm}
    \begin{tabular}{c c}
    \hline
         Type&Count  \\\hline
         Pronouns&31.96  \\
         Mention-Pronoun pairs&13.49\\
         \hline
    \end{tabular}
    \caption{count of pronouns and mention-pronoun pairs}
    \label{tab1}
\end{table}

To capture the feature introduced by pronouns, we propose a novel Coref-aware Doc-level Relation Extraction based on Graph Inference Network (CorefDRE), a fine-tuned Coref-aware approach that instructs the model directly to learn the coreference by mention-pronoun clustering. Intuitively, we propose a mention-pronoun coreference resolution, utilizing NeuralCoref, an extension to the Spacy, to extract the pronouns for each mention in the document, and using BERT to calculate the affinity of each mention-pronoun pair, to reduce the noise brought by pronouns. According to the mention-pronoun pairs and the affinities, the Mention-Pronoun Affinity Graph (MPAG) is constructed, which is a heterogeneous graph with pronoun node, mention node, and three types of edge. After that, we introduce GCN on MPAG to get the representation for each mention and pronoun. We further propose the noise suppression mechanism to merge the pronoun into MPAG, aggregated to isomorphic Entity Graph (EG), and obtain the final representation of node and edge in the graph.

Our contributions are summarized as follows: 
\begin{itemize}

\item We introduce a novel heterogeneous graph, Mention-Pronoun Affinity Graph (MPAG), including mention-pronoun pairs and corresponding affinity, to better model Doc-level Relation Extraction task.

\item We propose a mention-pronoun affinity model to calculate the affinity between mention and corresponding pronoun for coreference resolution and the noise suppression mechanism to merge the pronoun into MPAG through the weight of the mention-pronoun edge.

\item We conduct experiments on DocRED, DialogRE and MPDD dataset. Experimental results demonstrate the effectiveness of our CorefDRE model that achieves state-of-the-art performance.

\end{itemize}

\section{Problem Formulation}

Given a document $\mathrm{D}$ containing n sentence $\left\{\mathrm{s}_{1}, \mathrm{~s}_{2}, \ldots, \mathrm{s}_{\mathrm{n}}\right\}$ and a entity list $E=\left\{e_{1}, e_{2}, \ldots, e_{m}\right\}$, document-level relation extraction task is to extract the relation $r_{s, o}$ between the subject and object entity pair $\left(e_{s}, e_{o}\right)$. In the above definition, $\mathrm{s}_{\mathrm{i}}=\left\{w_{1}, w_{2}, \ldots, w_{k}\right\}$ means that $\mathrm{s}_{\mathrm{i}}$ consists of $k$ words, $\mathrm{e}_{\mathrm{i}}=\left\{m_{1}, m_{2}, \ldots, m_{g}\right\}$ means that there are $g$ mentions belong to $e_{i}$ and $m_{i}=\left\{w_{1}, w_{2}, \ldots, w_{l}\right\}$ means that $m_{i}$ consists of $l$ consecutive words.
The mention-pronoun pair, defined as $\left\{\left(m_{i}, p_{j}\right) \mid m_{i} \in E, p_{j} \in M_{p_{i}}\right\}$,
is used for coreference resolution to optimize relation extraction from document. $M_{p_{i}}=\left\{p_{1}, p_{2}, \ldots, p_{l i}\right\}$ is the pronouns set referring to $\mathrm{m}_{\mathrm{i}}$ and each pronoun $\mathrm{p}_{\mathrm{j}}$ can form a mention-pronoun pair with $\mathrm{m}_{\mathrm{i}}$, e.g., [$\left< Colette~de~Jouvebel, she \right>$ , $\left<Bel-Gazou, she \right>$] is shown in Figure~\ref{fig1}.

\section{Proposed Approach}

\begin{figure*}[ht]
    \centering
    \setlength{\abovecaptionskip}{0cm}
    \setlength{\belowcaptionskip}{-0.4cm}
    \vspace{-0.8cm}
    \includegraphics[width=6.6in]{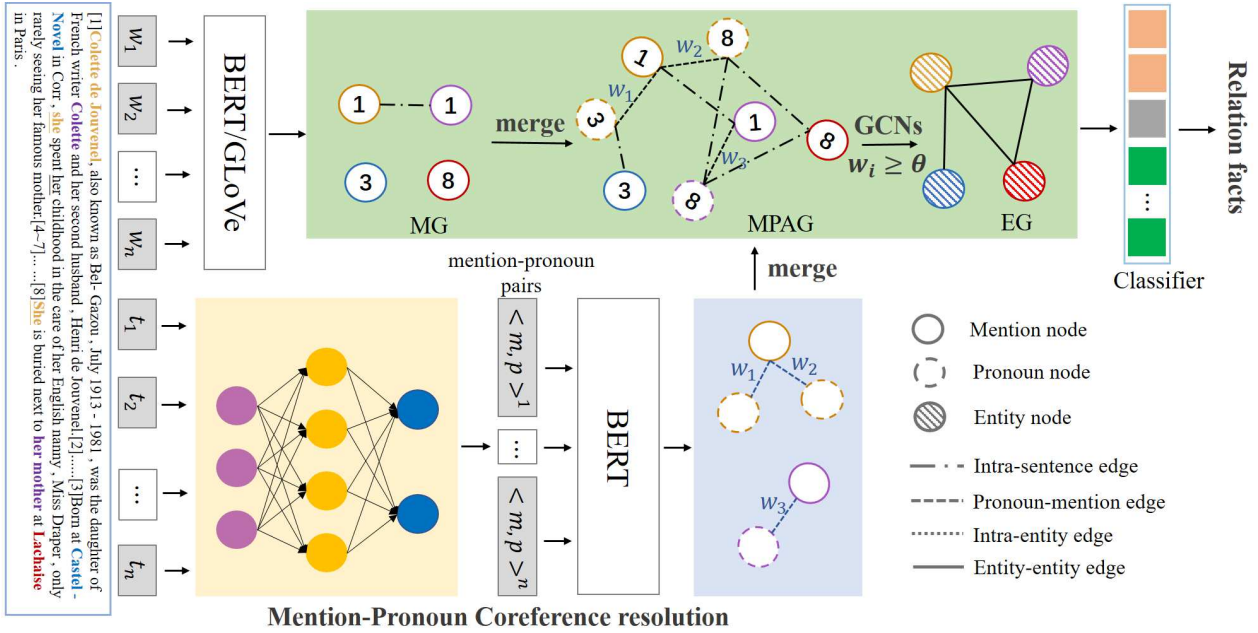}
    \caption{Architecture of our CorefDRE. First, the document is fed into encoder respectively, and then MPAG is constructed with pronoun nodes and mention nodes. Second, mention-pronoun coreference resolution takes use of contextualized representation and mention-pronoun pairs to calculate affinity. Third, merge the output of mention-pronoun coreference resolution to MPAG with noise suppression mechanism by applying GCN. Finally, the graph is transformed into EG, where the paths between entities are identified for reasoning. Different entities are drawn with colors, the squares represent the relationship between entity pairs, and the number in each circle is the sentence number.}
    \label{fig2}
\end{figure*}

To extract pronouns and relations from the document, we introduce CorefDRE to model the relations of graph nodes consisting  of pronoun, mention, and entity by constructing Mention-Pronoun Affinity Graph (Section 3.2) and mention-pronoun coreference resolution (Section 3.3), and then merge the coreference representation to MPAG through dynamically graph inference (Section 3.4), as is shown in  Figure~\ref{fig2}.

\subsection{Method Overview}

To motivate our approach, we perform the problem by learning a pronoun-mention graph representation, from which the derived graph can explicitly model the relation among pronouns and mentions to infer the entailment in the document. As illustrated in Figure~\ref{fig2}, we learn graph representation by rendering the following three steps: Firstly, we construct a mention graph and then dynamically merge the mention-pronoun pairs to conduct a mention-pronoun affinity graph.
Secondly, mention-pronoun coreference resolution is performed almost parallel to the mention graph and then as the basis of generating MPAG in step 1.
Finally, to denoise the relation classifier, MPAG makes the decision whether the mention-pronoun pair, produced by mention-pronoun coreference resolution in step 2, is merged into the graph by noise suppression mechanism in graph inference or not.

\subsection{Mention-Pronoun Affinity Graph}

Pronouns are extremely vital to extract the relation facts between two entities crossing multi-sentence from a document. Therefore, we identify the representations that refer to the same mention and cluster them together as mention-pronoun pairs. Then, we construct our Mention-Pronoun Affinity Graph (MPAG) according to text and mention-pronoun pairs. MPAG has two types of nodes and three types of edges: 

Mention Node: Each mention in the graph corresponds to a mention node, which is defined by the concatenation of mention semantic and type representation $t_{m} \in R^{dt}$. Thus, the representation of $m_{i}=\left[avg_{w_{k} \in m_{i}}\left(h_{k}\right) ; t_{m}\right]$ is referred to as mention node, where $avg_{w_{k} \in m_{i}}\left(h_{k}\right)$ is the average representation of mention contained words encoded by encoder.

Pronoun Node: Each pronoun (like $it$, $his$, $she$) refers to the special mention in the document corresponding to a pronoun node, which has a type representation $t_{p} \in R^{dp}$.

Intra-Entity Edge:  If two mention nodes refer to the same entity, there is an intra-entity edge between them. The edge can model the interaction among different mentions of the same entity and establish the interaction among the mentions of cross sentences.

Intra-Sentence Edge: If mention nodes or pronoun nodes appear in the same sentence, there is an intra-sentence edge between them. The edge can model the interaction among different mentions and pronouns of the same sentence and establish the interaction among the mentions and pronouns referring to different entities.

Mention-Pronoun Edge: mention-pronoun edge is established according to mention-pronoun pairs. For pair $\left<m, p\right>$, there is a mention-pronoun edge between mention node $m$ and pronoun node $p$. The can strengthen the interaction of semantic information among sentences.

What needs to be emphasized is that other interactions are implicitly contained in MG. 
To initialize the graph MG, we follow the GAIN proposed by \cite{zeng2020double} and dynamically construct MPAG by applying Graph Convolution Network \cite{kipf2017semi} to convolute the heterogeneous graph. Given node m forward-pass update for the ($l$+$1$)$th$ layer, the heterogeneous graph convolutional operation is defined as follows:
 
\begin{equation}
n_{i}^{l+1}=\sigma\left(\sum_{e \in E} \sum_{j \in N} \frac{1}{|N|} W_{e}^{l} n_{j}^{l}+b_{e}^{l}\right)
\label{eql1}
\end{equation}

where $\sigma(.)$ is the activation function. $E$ denotes the set of different edges, $N$ denotes the set of different neighbors of node $n$ and $W_{e}^{l}$, $b_{e}^{l} \in R^{d \times d}$ are trainable parameters.

To cover features of all levels, the final representation of node $\mathrm{n}_{i}$ can be concatenated from each layer:

\begin{equation}
n_{i}=\left[n_{i}^{0} ; n_{i}^{1} ; \ldots ; n_{i}^{N}\right]
\label{eql2}
\end{equation}

where $n_{i}^{0}$ is the initial representation of node $n_{i}$ and is formed by the document representation from encoder.

\subsection{Mention-Pronoun Coreference Resolution}

Mention-pronoun coreference resolution is the task to identify the pronouns that refer to the same mention and cluster the mention-pronoun pairs together as coreference clusters. For instance, ``$Niko$ $Nirvi$ ($born$ $in$ $1960$) $is$ $a$ $long$-$term$~$major$~$icon$~$in$~$the$~$Finnish$~$gaming$~$world$.~$He$~$is$ $well$~$known$~$for$~$writing$~$computer$~$game$~$reviews$~$since$~$the$ $1980s$~$in$~$Mikro$~$Bitti$, $Nirvi$~$began$~$his$~$career$~$as$~$a$ $game$~$reviewer$~$in$~$1986$~$on$~$pages$~$of$~$Mikro$~$Bitti$.", we can obtain a mention-pronoun pair clusters simply, e.g., [$\left<Niko~Nirvi, He\right>$,$\left<Niko~Nirvi, his\right>$,\dots,$\left<Nirvi, He\right>$], by the mention-pronoun coreference resolution. 

Mention-pronoun pair may lead to noise for Doc-level relation extraction because of the complex semantic in the document, we take advantage of BERT to measure the affinity of the $ \left<mention,pronoun \right>$ pair relationship. For each pair $ \left<mention, pronoun \right>$, we concatenate the context of $ \left<mention, pronoun \right>$ pair as input and produce a single affinity scalar for every pair when constructing MPAG. The input form of tokens is as follows:

\begin{equation}
\begin{gathered}
{[CLS] \left<\text { Mention }\right>[SEP]\left<\text { Pronoun }\right>[SEP]} \\
where \left<\star\right>:=c_{l}[S T A R T] \star[E N D] c_{r}
\end{gathered}
\label{eql3}
\end{equation}

where $\star$ is mention tokens or pronoun tokens and $c_{l}$ , $c_{r}$ represent the text on left and right of "$\star$" respectively. The $\left\lbrack {START} \right\rbrack$ and $\left\lbrack {END} \right\rbrack$ are special tokens fine-tuned that indicate the start and end of "$\star$" in the context respectively. 

We make affinity symmetric by averaging the representation of $\left< mention, pronoun \right>$ and $\left< pronoun,mention \right>$ to improve the representation. And then the affinity of the mention-pronoun pair is calculated by the enhanced representation of pairs and passed into a linear layer with sigmoid activation. For instance, the affinity between mention pair $\left< mention, pronoun \right>$ is set 1, which is a strong signal for the fusion of MPAG. To calculate the affinity between the mention-pronoun pair accurately, we design subtly the positive sampling and negative sampling to train the affinity calculation. We screen out 300 positive samples $D_{p}$ from the data $D$ obtained by coreference resolution and replace the mention $m$ of the positive sample with other mentions $m^{\prime}$ randomly. To train the model that calculates the affinity of $\left<mention, pronoun \right>$, we minimize the following triplet max-margin loss when training.

\begin{equation}
L_\varphi=\ \sum_{p_+,m\epsilon P^+\ }\sum_{p_-,m\epsilon P^-\ } l\left(m,p_+,p_-\right)
\label{eql4}
\end{equation}

\begin{equation}
{l}(g,p,n)=\left[aff(g, n)^{2}-(1-aff(p, n))^{2}\right]_{+}
\label{eql5}
\end{equation}

where $m$ and $p$ are mention and pronoun in mention-pronoun pair $\left<m, p\right>$ and $aff\left(m,p \right)$ is the affinity between $m$ and $p$. The $g$, $p$, $n$ in formula~\ref{eql5} are mention, negative pronoun and positive pronoun referring to mention.

\subsection{Graph Inference}

Inspired by \cite{zeng2020double}, we predict relation facts between entity pairs by reasoning on Entity Graph (EG), which is transformed from MPAG. The dynamic process of merging MPAG to EG, defined as noise suppression mechanism, is divided into three steps: 

Step 1: pronoun nodes that refer to the same mention are merged with the mention node to form a new mention node. Note that if the affinity between the mention-pronoun pair is less than the threshold $\theta$, the pronoun does not participate in the merging process so that noise is depressed simply. For the 
$i$-$th$ mention node merged from $N$ pronoun nodes, it is represented by concatenating the mention and the average of its $N$ pronoun node representations, and mention node representation is defined as:

\begin{equation}
{m}_{i}=\bar{m_{i}} \oplus  \frac{1}{~N} \sum_{n} aff_{n} p_{n}\left({aff}_{n} \geq \theta\right)
\label{eql6}
\end{equation}

where $\bar{\mathrm{m}_{\mathrm{i}}}$ denotes the mention representation, $p_{n}$ is the $n$-$th$ referred to the mention $m_{i}$ and $aff_{n}$ is the affinity of $\left< mention,pronoun \right>$ pair and $\oplus$ denotes concat operation.

Step 2: mention nodes that refer to the same entity are merged to an entity node in EG. For the $i$-$th$ entity node merged for $N$ mention nodes, it is represented by the average of its $N$ mention node representation:

\begin{equation}
e_{i}=\frac{1}{N} \sum_{n} m_{n}
\label{eql7}
\end{equation}

Step 3: intra-entity edges between the mentions, which refers to the same two entities, is merged as the edge in EG. The directed edge between entity nodes $e_{i}$ and $e_{j}$ in EG is defined as:

\begin{equation}
\text edge_{ij}=\sigma\left(W_{q}\left[e_{i} ; e_{j}\right]\right)+b_{q}
\label{eql8}
\end{equation}

where $W_{q}$ and $b_{q}$ are trainable parameters and $\sigma$ is an activation function (e.g., ReLU).

We model the potential reasoning clue between the entity nodes in EG through the path between the entity nodes. Based on the representation of the edge, $h-hop$ path between entity nodes $e_{s}$ and $e_{o}$ is defined as: 

\begin{equation}
\begin{gathered}
\mathrm{P}_{s, o}^{k}=[edge_{s, i_{1}} ; \ldots ;edge_{i_{h-2}, i_{h-1}} ;edge_{i_{h-1}, o} ; \\
 edge_{o, i_{h-1}} ; \ldots ; edge_{i_{2}, i_{1}} ; edge_{i_{1}, s} ]
\end{gathered}
\label{eql9}
\end{equation}

where $edge_{m,i_{n}~}$ stands for the edge between the $m^{th}$ and the $(n-1)^{th}$ intermediate node, and the two-hop path, according to our experiment, is selected in our model because of balancing the precision and performance. Since there are multiple paths between two entity nodes, an attention mechanism is introduced to fuse the path information and pay more attention to the strong path. Path information of the entity in EG is defined as:

\begin{equation}
s_{i}=\sigma\left(\left[e_{s} ; e_{o}\right] \cdot W_{l} \cdot p_{s, o}^{i}\right)
\label{eql10}
\end{equation}

\begin{equation}
\alpha_{i}=\frac{e^{s_{i}}}{\sum_{j} e^{s_{j}}}
\label{eql11}
\end{equation}

\begin{equation}
p_{s, o}=\sum_{i} \alpha_{i} p_{s, o}^{i}
\label{eql12}
\end{equation}

where $\alpha_{i}$ is the attention weight for $i^{th}$ path and $\sigma$ is an activation function (e.g., ReLU). 

According to the fusion of MPAG and mention-pronoun coreference, EG is dynamically constructed, which is an isomorphic graph and is converted from MPAG. The node and edge in EG can be represented by fusing the representation of mention and pronoun nodes, and the relationship between entity nodes can be predicted by the path inference. To identify the relationship of entity pair $\left< e_{s},e_{o} \right>$, we concatenate the following representations as $I_{s,o}$ and into the MLP: 

\begin{equation}
I_{s, 0}=\left[e_{s} ; e_{t} ;\left|e_{s}-e_{o}\right| ; e_{s} \odot e_{o} ; p_{s, o}\right]
\label{eql13}
\end{equation}

where $e_{s}$ and $e_{o}$ are the representation of subject and object entity in EG and $p_{s, 0}$ is the comprehensive inferential path information. Our loss function uses binary cross entropy to train our model:

\begin{equation}
L=-\sum_{D \in S} \sum_{s \neq o} \sum_{r \in R} \text { $CrossEntropy$ }\left(P_{r}\left(I_{s, o}\right), \overline{y_{r}}\left(I_{s, o}\right)\right)
\label{eql14}
\end{equation}

where $S$ denotes the whole corpus, $\overline{y_{r}}\left(I_{s, o}\right)$ refers to ground truth.
\begin{table*}[htbp]
\centering
\setlength{\belowcaptionskip}{-0.4cm}
\begin{tabular}{lllll} 
\toprule
\textbf{Model} & \multicolumn{2}{c}{\textbf{Dev}}&\multicolumn{2}{c}{\textbf{Test}}   \\
\cline{2-5} 
   &    Ign F1 & F1 & Ign F1 & F1  \\
  \midrule
  CNN$^{*}$~\cite{yao2019docred} &  41.58 & 43.45 & 40.33 & 42.46   \\
  BiLSTM$^{*}$~\cite{yao2019docred} &  48.87 & 50.94 & 48.78 & 51.06   \\
  ConText-Aware$^{*}$~\cite{yao2019docred} &  48.94 & 51.09 & 48.40 & 50.70   \\
  GAIN-GloVe$^{*}$~\cite{zeng2020double} &  53.03 & 55.29 & 52.66 & 55.08   \\
  \midrule
  \textbf{CorefDRE- GloVe} &  \textbf{55.01}& \textbf{57.33} & \textbf{54.37} & \textbf{56.74}   \\
  \midrule
  \midrule
  {$\rm BERT_{base}$$^{*}$} ~\cite{wang2019fine} &- &54.16 &- &53.20 \\
  {$\rm CorefBERT_{base}$$^{*}$} ~\cite{ye2020coreferential} & 55.32 & 57.51 & 54.54 & 56.96 \\
  {$\rm DocuNet$-$\rm BERT_{base}$$^{*}$} ~\cite{zhang2021document} & 59.86 & 61.83 & 59.93 & 61.86 \\
  {$\rm GAIN$-$\rm BERT_{base}$$^{*}$} ~\cite{zeng2020double} & 59.15 & 61.22 & 59.00 & 59.05 \\
  \midrule 
  \textbf{CorefDRE- BERT$_{base}$} &  \textbf{60.85}& \textbf{63.06} & \textbf{60.78} &\textbf{60.82}   \\
  \bottomrule
  \end{tabular}
  \caption{Performance on DocRED. Models above the first double line do not use pre-trained models. Results with * are reported in their original papers. Ign F1 refers to excluding the relational facts shared by the training and dev/test sets.}
  \label{tab2}
\end{table*}
\section{Experiments}
\subsection{Experimental settings}

\textbf{DocRED} \cite{yao2019docred}: More than 40.7$\%$ of the relation facts require reasoning over multiple sentences. \textbf{DialogRE} \cite{yu2020dialogue}: 95.6$\%$ of relational triples can be inferred through multiple sentences, where pronouns are extensively used. \textbf{MPDD} \cite{chen2020mpdd}: A publicly available Chinese dialogue dataset have both the emotion and interpersonal relation labels, which also have a mass of pronouns.
To learn an effective representation for document and capture the context of each mention, Following \cite{yao2019docred}’s work, for each word, we concatenate its word embedding, entity type embedding and entity id embedding. And then we feed all the word representations into Glove/BERT to get the representation of the document. We extract the relation between pronoun and mention based on Huggingface’s NeuralCoref and use BERT to pretrain the affinity for mention-pronoun pair. We 
use GloVe or uncased BERT base as the encoder, and 2 layers of GCN to encode the MPAG and EG. Our model is optimized with AdamW \cite{loshchilov2017decoupled} and set the dropout rate of GCN to 0.6, learning rate to 0.001.
\subsection{Baseline Models}
We use the following models as baselines.

\textbf{CNN $\&$ BiLSTM}: \cite{yao2019docred} proposed CNN and BiLSTM to encode the document into a sequence of the hidden state vectors. \textbf{Context-Aware}: \cite{yao2019docred} also proposed LSTM to encode the document and attention mechanism to fuse contextual information for predicting. \textbf{CorefBERT}: a pre-trained model was proposed by \cite{ye2020coreferential} for word embedding. \textbf{DocuNet-BERT}: \cite{zhang2021document} proposed a U-shaped segmentation module to capture global information among relational triples.
\textbf{GAIN-GloVe/GAIN-BERT}: \cite{zeng2020double} proposed GAIN, which designed mention graph and entity graph to predict target relations, and make use of GloVe or BERT for word embedding, GCN for representation of the graph. 

\subsection{Main Result}

\begin{table}
\centering
\setlength{\belowcaptionskip}{-0.4cm}
\begin{tabular}{lllll} 
\toprule
\textbf{Model} & \multicolumn{2}{c}{\textbf{Dev}}&\multicolumn{2}{c}{\textbf{Test}}   \\
\cline{2-5} 
   &    Ign F1 & F1 & Ign F1 & F1  \\
  \midrule
  CorefDRE-GloVe &  55.01& 57.33 & 54.37 & 56.74  \\
  -pronoun node &  53.12 & 55.37 & 52.71 & 55.18   \\
  -weighted edge &  53.57 & 55.35 & 53.02 & 55.26   \\
  \midrule
  CorefDRE-BERT$_{base}$ &  60.85 & 63.06 & 60.78 & 60.82   \\
  -pronoun node &  59.21 & 61.25 & 59.03 & 59.12   \\
  -weighted edge &  59.57 & 61.67 & 59.43 & 59.64   \\
  \bottomrule
  \end{tabular}
  \caption{Performance of CorefDRE with different embeddings and submodules.}
  \label{tab3}
\end{table}

\begin{table}
\centering
\setlength{\belowcaptionskip}{-0.4cm}
\begin{tabular}{lcc}
\hline
    \textbf{Model}  & \textbf{ F1-DialogRE} & \textbf{Acc-MPDD} \\
\hline
CNN\cite{yu2020dialogue}       & 46.1  & -     \\
BERT\cite{long2021consistent}       & 60.6  & 31.0      \\
GAIN\cite{long2021consistent}     & 69.8  & 42.2   \\
CoIn\cite{long2021consistent}     & 71.1  & 46.5   \\
\hline
\textbf{CorefDRE}   & \textbf{71.4}  & \textbf{46.7}    \\
\hline
\end{tabular}
\caption{Performance on the datasets DialogRE and MPDD}
\label{tab4}
\end{table}
We compare our CorefDRE model with other baselines on the DocRED dataset. The results are shown in Table~\ref{tab2}. We use F1 and Ign F1 as evaluation indicators to evaluate the effect of models.
Compared with the models based on GloVe, CorefDRE outperforms strong baselines by 1.7$\sim$2.0 F1 scores on the development set and test set. Compared with the models on BERT-base, CorefDRE outperforms strong baselines by 1.6$\sim$1.8. These results suggest that the mention-pronoun affinity graph can capture the interaction relationship of multi-sentences for better doc-level relation extraction. Although we only conduct the experiments on DocRED, DialogRE and MDPP shown in Table~\ref{tab4}, our model, obviously, is fit to others since pronouns is the essential grammar and syntax of the natural language.

\subsection{Ablation Study}

To verify the effectiveness of different modules in CorefDRE, we further analyze our model and the results of the ablation study shown in Table~\ref{tab3}.
First, we remove the affinity between the pronoun node and mention node. We set the weight of the mention-pronoun edge directly to 1 and merge all the pronoun nodes with the corresponding mention node when generating EG. Without the weight between pronoun node and mention node, the performance of CorefDRE-GloVe/CorefDRE-BERT$_{base}$ sharply drops by 1.39 F1 on the development set. This drop shows that the affinity between pronoun node and mention node plays a vital role in suppressing the noise caused by unsuitable mention-pronoun pairs.

Next, we remove the pronoun nodes. Specifically, we convert the MPAG into the MG proposed by \cite{zeng2020double}. Without pronoun nodes, the result drops by 1.81 F1 on the development set. This suggests that the pronoun nodes can capture richer information that mention node and document cannot capture effectively. 

\subsection{Case Study}
\begin{figure*}[ht]
    \centering
    \setlength{\abovecaptionskip}{0.4cm}
    \setlength{\belowcaptionskip}{-0.4cm}
    \vspace{-0.8cm}
    \includegraphics[width=6.6in]{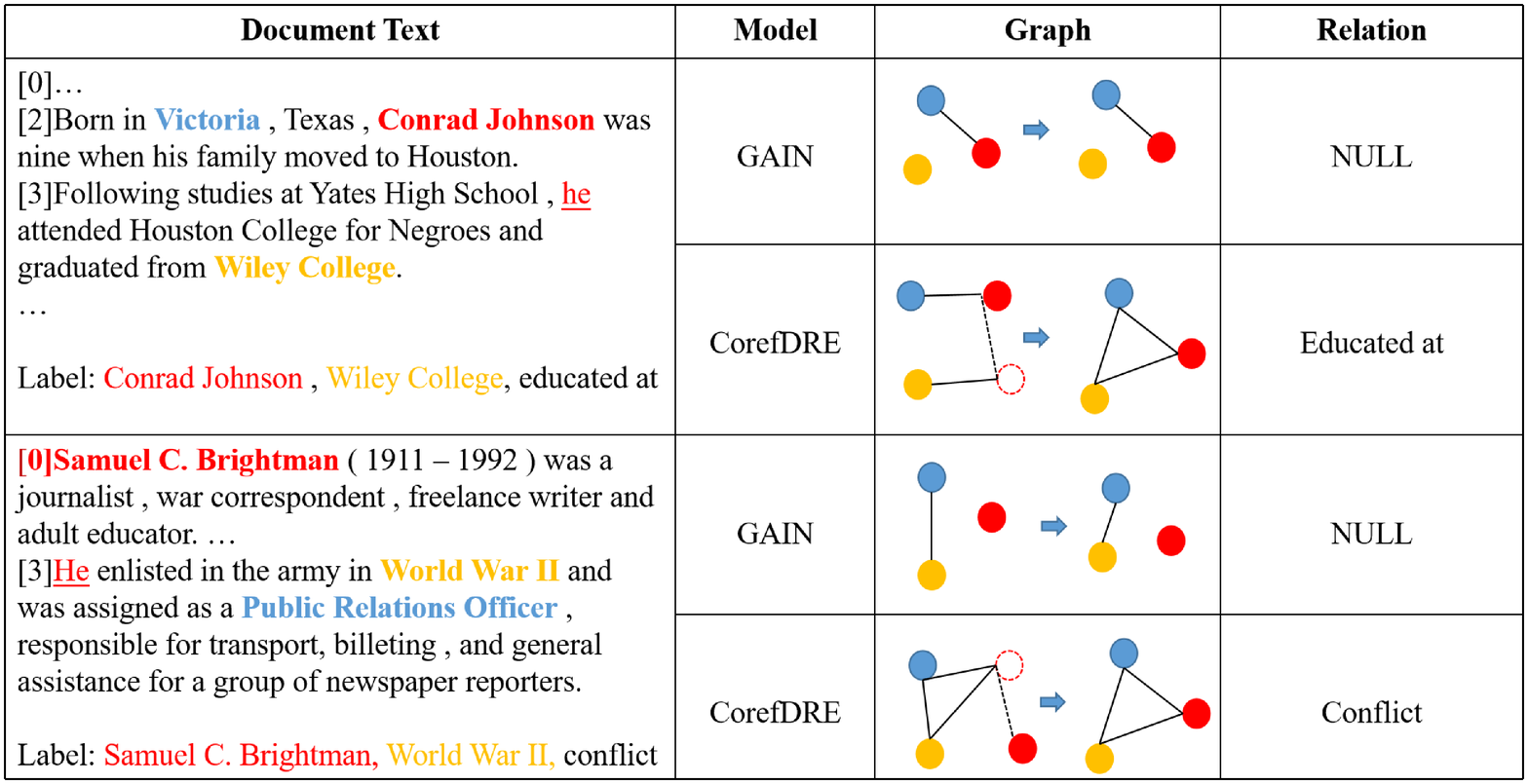}
    \caption{Case Study on our CorefDRE model and baseline model. The graph indicates our model based on MPAG can extract the relations while other models, GAIN, have not the inference.}
    \label{fig3}
\end{figure*}
Figure \ref{fig3} illustrates the case study of our CorefDRE compared with our baseline. As is shown, GAIN can not predict the relation of entity pairs $\left<Conrad~Johnson, Wiley~College\right>$ and $\left<Samuel~C.~Brightman, World~War~II\right>$, while CorefDRE can predict the relation between $Conrad$ $Johnson$ and $Wiley$ $College$ is $educated$ $at$ and the relation between $Samuel C.$ $Brightman$ and $World$ $War$ $II$ is $conflict$, because pronoun nodes $he$ and $He$ can connect the entity pair $\left<Conrad~Johnson,Wiley~College\right>$ and $\left<Samuel~C.~Brightman, World~War~II\right>$ respectively. We observe that relation extraction among those entities need pronouns to connect them across sentences. The observation proves the effectiveness of our model.

\section{Related work}

Relation Extraction is to extract relation facts from a given text, while early research focus is mainly on predicting relation fact between two entities within a sentence \cite{zeng2015distant,wang2016relation,zhang2017position,zhao2021modeling,liu2021attention,guo2021learning,shang2022pattern}. These approaches include sequence-based methods, graph-based methods and pre-training methods, which can tackle sentence-level RE effectively, and the dataset contains very limited relation types and entity types. However, large amounts of relation facts only can be extracted through multiple sentences. 

\textbf{Document-level relation extraction}.  Researchers extend sentence-level to document-level RE \cite{christopoulou2019connecting,wang2020global,zhang2021document} and explore two trends. The first is the sequence-based method that uses the pre-trained model to get the contextual representation of each word in a document, which directly uses the pre-trained model to obtain the relationship between entities \cite{ye2020coreferential}. These methods adopt transformers to model long-distance dependencies implicitly and get the entities embedding, and feed them into a classifier to get relation labels. But the sequence-based method cannot capture enough semantic relations when the document length is out of the capability of the encoder at a time. Another trend is the graph-based method that constructs graphs according to documents, which can model entity structure more intuitively  \cite{sahu2019inter,zhu2019graph,zeng2020double}. These methods take advantage of LSTM or BERT to encode the input documents and output the representation of entities to the GCNs to update the representation and then feed them into the classifier to get relation labels.

\textbf{Coreference dependency relation reasoning}. Some previous efforts on document-level RE introducing coreference dependency for multi-hop inference are useful for solving multi-hop reasoning. Previous works \cite{zhu2019graph,sahu2019inter,fu2021end} have shown that graph-based coreference resolution is obviously beneficial to construct dependencies among mentions for relation reasoning. \cite{zeng2021sire} proposed intra-and-inter-sentential reasoning based on R-GCN to model multiple paths by covering all cases of logical reasoning chains in the graph. \cite{xu2021discriminative} introduced a reconstructor to rebuild the graph reasoning paths to guide the relation inference by multiple reasoning skills including coreference and entity bridge. However, none of the above methods model the influence of pronouns on relation extraction and reasoning directly. Our CorefDRE model deals with the problem by introducing a novel heterogeneous graph with mention-pronoun coreference resolution and noise suppression mechanism.  

\section{Conclusion and Future Work}

We propose the CorefDRE which features three novel skills: coref-aware heterogeneous graph, mention-pronoun coreference resolution, and noise suppression mechanism. Based on the proposed method, the model can extract Doc-level entity pair relation more effectively due to the richer pronoun bridging representation. Experiments demonstrate that our CorefDRE outperforms previous models significantly and is orthogonal to pretrained language models. However, there are still some problems not completely solved, where the noise produced by pronouns hinders our model to improve performance. In the future, we will explore other methods to construct mention-pronoun pairs to optimize CorefDRE.

{\small
\bibliographystyle{named}
\clearpage
\bibliography{ijcai22}
}
\end{document}